\documentclass{article}

\usepackage[final,nonatbib]{nips_2017}

\usepackage[utf8]{inputenc} 
\usepackage[T1]{fontenc}    
\usepackage{hyperref}       
\usepackage{url}            
\usepackage{booktabs}       
\usepackage{amsfonts}       
\usepackage{nicefrac}       
\usepackage{microtype}      

\usepackage{tabularx}
\usepackage{amsmath,amssymb,amsthm}
\usepackage{graphicx}
\usepackage{subcaption}
\usepackage{listings}
\usepackage{upquote}
\usepackage{wrapfig}
\usepackage{lipsum}
\usepackage{array}
\usepackage{multicol}
\usepackage{multirow}
\usepackage{url}
\usepackage{color}
\usepackage{wrapfig,lipsum,booktabs}

\title{Trimming and Improving Skip-thought Vectors}

\author{
Shuai Tang$^1$ \hspace{0.5cm} Hailin Jin$^2$ \hspace{0.5cm} Chen Fang$^2$ \hspace{0.5cm} Zhaowen Wang$^2$ \hspace{0.5cm} Virginia R. de Sa$^1$ \\
Department of Cognitive Science, UC San Diego $^1$ \\
Adobe Research $^2$ \\
\texttt{\{shuaitang93,desa\}@ucsd.edu}, \texttt{\{hljin, cfang, zhawang\}@adobe.com}
}

\begin{document}

\maketitle

\begin{abstract}
The skip-thought model has been proven to be effective at learning sentence representations and capturing sentence semantics. In this paper, we propose a suite of techniques to trim and improve it. First, we validate a hypothesis that, given a current sentence, inferring the previous and inferring the next sentence provide similar supervision power, therefore only one decoder for predicting the next sentence is preserved in our trimmed skip-thought model. Second, we present a connection layer between encoder and decoder to help the model to generalize better on semantic relatedness tasks. Third, we found that a good word embedding initialization is also essential for learning better sentence representations. We train our model unsupervised on a large corpus with contiguous sentences, and then evaluate the trained model on 7 supervised tasks, which includes semantic relatedness, paraphrase detection, and text classification benchmarks. We empirically show that, our proposed model is a faster, lighter-weight and equally powerful alternative to the original skip-thought model.
\end{abstract}

\section{Introduction}
Learning distributed sentence representations is an important and hard topic in both the deep learning and natural language processing communities, since it requires machines to encode a sentence with potential unlimited language content into a fixed-dimension vector filled with continuous values. We are interested in learning to build a distributed sentence encoder in an unsupervised fashion by exploring the structure and relationship in a large unlabelled corpus. Since humans understand sentences by composing from the meanings of the words, we define that learning a sentence encoder should be composed of two essential components, which are learning distributed word representations, and learning how to compose a sentence representation from the representations of words in the given sentence. 

With the development of deep learning techniques, recurrent neural networks (RNNs) \cite{Elman1990FindingSI,Hochreiter1997LongSM,Chung2014EmpiricalEO} have shown encouraging results on natural language processing (NLP) tasks, and become the dominant methods in processing sequential data. \cite{Sutskever2014SequenceTS} proposed LSTM-based sequence to sequence learning (seq2seq) model for machine translation. Later \cite{Dai2015SemisupervisedSL} applied the seq2seq model for unsupervised representation learning on language, and then finetuned the model for supervised tasks. The seq2seq model can be jointly trained to learn the word representation and the composition function on word representations, also it shows encouraging idea that knowledge learned from unsupervised training could be transferred to help other related supervised tasks.

\cite{Kiros2015SkipThoughtV} proposed the skip-thought model, which is an encoder-decoder model for unsupervised distributed sentence representation learning. The paper exploits the semantic similarity within a tuple of adjacent sentences as a supervision, and successfully built a generic, distributed sentence encoder. Rather than applying the conventional autoencoder model, the skip-thought model tries to reconstruct the surrounding 2 sentences instead of itself. The learned sentence representation encoder outperforms previous unsupervised pretrained models on the evaluation tasks with no finetuning, and the results are comparable to the models which were trained directly on the datasets in a supervised fashion. 

In this paper, We aim to trim and improve the original skip-thought model by three techniques. First, given the neighborhood hypothesis first proposed in \cite{Tang2017Rethinking}, we directly abandon one of the decoders in the skip-thought model, and leave only one encoder and one decoder for learning from inferring the next sentence given the current one. Second, we replace the plain connection used between the encoder and decoder with the \emph{Average+Max Connection}, which is a non-linear non-parametric feature engineering method proposed by \cite{Chen2016EnhancingAC} for Stanford Natural Language Inference (SNLI) \cite{Bowman2015ALA} challenge, and enhances the model to capture more complex interactions among the hidden states. Third, a good initialization for word embeddings boosts the transferability of the model trained in unsupervised fashion, which may raise the importance of the word embeddings in unsupervised learning algorithms. In addition, we show that increasing the dimension of the encoder improves the performance of our proposed model, but still keeps the number of parameters much smaller than the original skip-thought model. Detailed description of our model is described in Section \ref{approach}.

\begin{figure*}[th]
\centering
\begin{subfigure}[t]{0.28\textwidth}
\includegraphics[width=\linewidth]{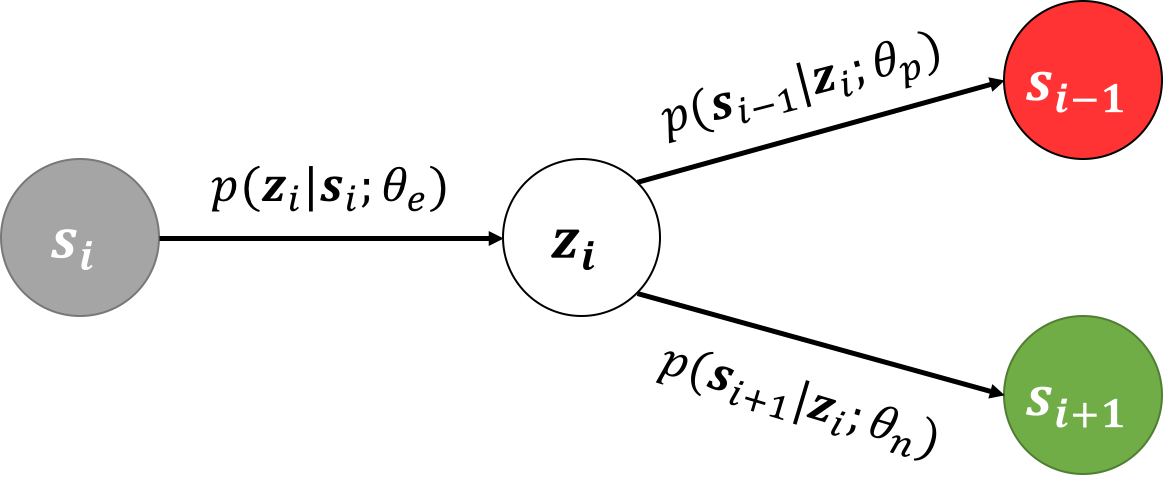}
\caption{Skip-thought}
\label{simpleskip}
\end{subfigure}
~
\begin{subfigure}[t]{0.28\textwidth}
\includegraphics[width=\linewidth]{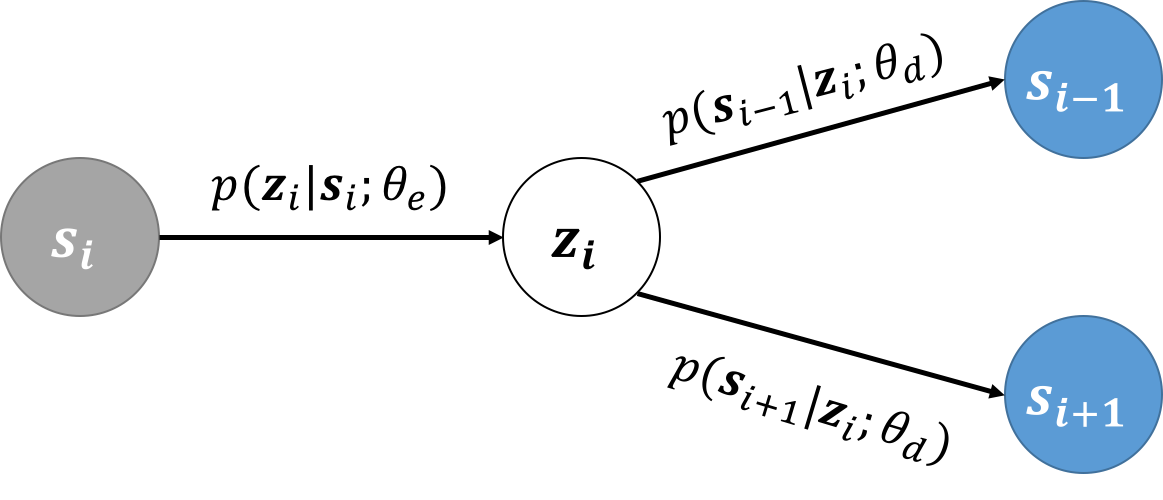}
\caption{Neighborhood Hypothesis}
\label{simpleneighbor}
\end{subfigure}
~
\begin{subfigure}[t]{0.33\textwidth}
\includegraphics[width=\linewidth]{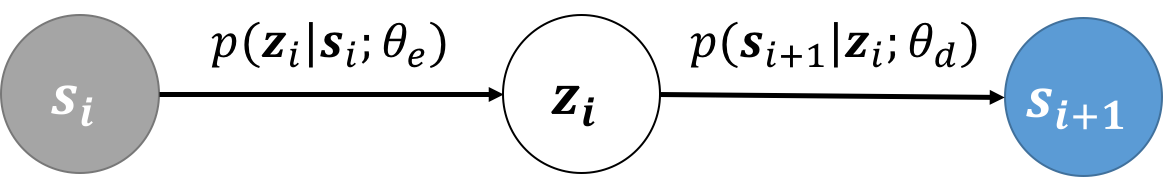}
\caption{Trimmed Skip-thought}
\label{trimmedskip}
\end{subfigure}
\caption{The comparison of the previously proposed skip-thought model \cite{Kiros2015SkipThoughtV}, and our proposed trimmed skip-thought model. Compared to the skip-thought model, our model only needs to reconstruct the next sentence per sample during training, which accelerates the training, and results in fewer parameters. Better view in color.}
\label{simplemodel}
\end{figure*}

\section{Approach}
\label{approach}
In this section, we present the trimmed skip-thought model. It includes a few simple yet effective modifications from the previously proposed skip-thought model \cite{Kiros2015SkipThoughtV}. We first briefly introduce the skip-thought model, and then discuss how to explicitly modify it by incorporating our proposed 3 techniques.

\subsection{Skip-thought Model}
In skip-thought model, given a sentence tuple $(s_{i-1},s_i,s_{i+1})$, the encoder computes a fixed-dimension vector as the representation $\mathbf{z}_i$ for the sentence $s_i$, which learns a distribution $p(\mathbf{z}_i|s_i;\theta_e)$, where $\theta_e$ refers to the set of parameters in the encoder. Then, conditioned on the representation $\mathbf{z}_i$, two separated decoders are applied to reconstruct the previous sentence $s_{i-1}$, and the next sentence $s_{i+1}$, respectively. We call them previous decoder $p(s_{i-1}|\mathbf{z}_i;\theta_{p})$ and next decoder $p(s_{i+1}|\mathbf{z}_i;\theta_{n})$, where $\theta_\cdot$ denotes the set of parameters in each decoder. An illustration is shown in Figure \ref{simpleskip}.

Since the two conditional distributions learned from the decoders are parameterized independently, they implicitly utilize the sentence order information within the sentence tuple. Intuitively, given the current sentence $s_i$, inferring the previous sentence $s_{i-1}$ is considered to be different from inferring the next sentence $s_{i+1}$.

\textbf{Encoder: } The encoder is a recurrent neural network, which is composed of bi-directional gated recurrent unit (GRU) \cite{Cho2014LearningPR}, or uni-directional GRU. Suppose sentence $s_i$ contains $N$ words, which are $w_i^1,w_i^2,...,w_i^N$. At an arbitrary time step $t$, the encoder produces a hidden state $\mathbf{h}_i^t$, and we regard it as the representation for the previous subsequence through time $t$. At time $N$, the hidden state $\mathbf{h}_i^N$ represents the given sentence $s_i$, which is $\mathbf{z}_i$. 

\textbf{Decoder: } The decoder is a single-layer recurrent network with conditional GRU. Specifically, compared to GRU, it takes the sentence representation $\mathbf{z}_i$ as an additional input at each time step. The decoder needs to reconstruct the previous sentence $s_{i-1}$ and the next sentence $s_{i+1}$ given the representation $\mathbf{z}_i$. The computation flows for the GRU and the conditional GRU are presented in Table \ref{cgru}.

\begin{table*}[h]
\fontsize{9}{13}\selectfont
\begin{center}
\begin{tabular}{c | c}
GRU & Conditional GRU \\
$\begin{bmatrix}
\mathbf{m}^t\\
\mathbf{r}^t\end{bmatrix}=\sigma\left(\mathbf{W}_h\mathbf{h}^{t-1}+\mathbf{W}_x\mathbf{x}^t\right) $ & $\begin{bmatrix}
\mathbf{m}^t\\
\mathbf{r}^t
\end{bmatrix}=\sigma\left(\mathbf{W}_h\mathbf{h}^{t-1}+\mathbf{W}_x\mathbf{x}^t+\mathbf{W}_z\mathbf{z}_i\right)$ \\
$\hat{\mathbf{h}}^t=\tanh\left(\mathbf{W}\mathbf{x}^t+\mathbf{U}\left(\mathbf{r}^t\odot\mathbf{h}^{t-1}\right)\right)$ & $\hat{\mathbf{h}}^t=\tanh\left(\mathbf{W}\mathbf{x}^t+\mathbf{U}\left(\mathbf{r}^t\odot\mathbf{h}^{t-1}\right)+\mathbf{U}_z\mathbf{z}_i\right)$ \\
$\mathbf{h}^t=(1-\mathbf{m}^t)\odot\mathbf{h}^{t-1}+\mathbf{m}^t\odot\mathbf{\hat{\mathbf{h}}}^t$ & $\mathbf{h}^t=(1-\mathbf{m}^t)\odot\mathbf{h}^{t-1}+\mathbf{m}^t\odot\mathbf{\hat{\mathbf{h}}}^t$ 
\end{tabular}
\end{center}
\caption{Here presents the Gated Recurrent Unit (GRU) \cite{Chung2014EmpiricalEO} and Conditional GRU, omitting the subscript $i$. $\mathbf{x}^t$ is the embedding for the word $w_i^t$, and $\mathbf{z}_i$ is the vector representation for sentence $s_i$. $\mathbf{W_\cdot}$ and $\mathbf{U_\cdot}$ are the parameter matrices, and $\odot$ is the element-wise product.}
\label{cgru}
\end{table*}

\subsection{Trimming Skip-thought Model by Neighborhood Hypothesis}
The neighborhood hypothesis was first introduced in \cite{Tang2017Rethinking}, and it pointed out that given the current sentence, inferring the previous sentence and inferring the next sentence both provide same supervision power.

To incorporate the neighborhood hypothesis into the model, we need to modify the skip-thought model. Given $s_i$, we assume that inferring $s_{i-1}$ is the same as inferring $s_{i+1}$. If we define ${s_{i-1}, s_{i+1}}$ are two neighbors of $s_i$, then the inferring process can be denoted as $s_j\sim p(s|\mathbf{z}_i;\theta_{d})$, for any $j$ in the neighborhood of $s_i$. The conditional distribution learned from the decoder is parameterized by $\theta_d$. Figure \ref{simpleneighbor} illustrates the neighborhood hypothesis.

Furthermore, in our trimmed skip-thought model, for a given sentence $s_i$, the decoder needs to reconstruct the sentences in its neighborhood $\{s_{i-1},s_{i+1}\}$, which are two targets. We denote the inference process as $s_i \rightarrow \{s_{i-1},s_{i+1}\}$. For the next sentence $s_{i+1}$, the inference process is $s_{i+1} \rightarrow \{s_i,s_{i+2}\}$. In other words, for a given sentence pair $\{s_i,s_{i+1}\}$, the inference process includes $s_i \rightarrow s_{i+1}$ and $s_{i+1} \rightarrow s_i$. 

In the neighborhood hypothesis \cite{Tang2017Rethinking}, the model doesn't distinguish between the sentences in a neighborhood. As a result, an inference process that includes both $s_i \rightarrow s_{i+1}$ and $s_{i+1} \rightarrow s_i$ is equivalent to an inference process with only one of them. Thus, we define a \emph{trimmed skip-thought} model with only one target, presented in Figure \ref{trimmedskip}, and the target is always the next sentence. The objective at each time step is defined as the log-likelihood of the predicted word given the previous words, which is
\begin{align}
&\ell^t_{i,j}(\theta_e,\theta_d) = \log p(w_j^t|w_j^{<t},\mathbf{z}_i;\theta_e,\theta_d)
\end{align}
where $\theta_e$ is the set of parameters in the encoder, and $\theta_d$ is the set of parameters in the decoder. The objective function is summed across the whole training corpus, then the objective during training is
\begin{align}
&\max_{\theta_e,\theta_d}\sum_i\sum_t \ell^t_{i,i+1}(\theta_e,\theta_d) 
\end{align}

\subsection{Average+Max Connection}
In skip-thought models \cite{Kiros2015SkipThoughtV}, only the hidden state at the last time step produced from the RNN encoder is regarded as the vector representation for a given sentence, and serves as the conditional information for the decoder to reconstruct the adjacent 2 sentences. 

Recently, \cite{Bowman2015ALA} collected a large corpus, which is SNLI, for textual entailment recognition. Given a sentence pair including premise and hypothesis, the task is to classify the relationship of the sentence pair, entailment, contradiction or neutral. \cite{Parikh2016ADA} proposed to summarize the hidden states from all time steps computed from a RNN encoder as a sentence representation. While \cite{Chen2016EnhancingAC} proposed to concatenate the outputs from an average pooling function and a max pooling function, which both run over all time steps, to serve as a sentence representation, and showed a performance boost on the SNLI dataset. 

The concatenation of an average pooling and a max pooling function is actually a non-parametric composition function, and the computation load is negligible compare to heavy matrix multiplication. Also, the non-linearity of the max pooling function augments the average pooling function for building a representation that captures more complex composition of the context information. Given a sentence $s_i$, the encoder produces a set of hidden states $[\mathbf{h}_i^1;\mathbf{h}_i^2;...;\mathbf{h}_i^N]$, the composition function could be represented as $\mathbf{z}_i=\left[\frac{1}{N}\sum_{n=1}^N\mathbf{h}_i^n;\max_{n=1}^N\mathbf{h}_i^n\right]$.

Here, since our goal is to simplify and accelerate the skip-thought model, and also get comparable results on the evaluation tasks, we consider comparing the 2 different composition functions, which are the original one used in the skip-thought model \cite{Kiros2015SkipThoughtV}, denoted as \emph{Plain Connection}, and the function proposed by \cite{Chen2016EnhancingAC}, denoted as \emph{Average+Max Connection}. We hypothesize that the composition function by concatenating two pooling functions will help the model perform better on tasks that involve judging the relationship of a sentence pair, while it is hard to say if the model would benefit from it on the classification benchmark. We will discuss the results in Section \ref{QE}.

\subsection{Word Embeddings Initialization}
Distributed word embedding matters in the deep learning models that deal with NLP-related tasks. The proposed training methods, such as continuous bag-of-words and skip-gram \cite{Mikolov2013EfficientEO}, always serves as strong baseline models for the supervised tasks in NLP. The pretrained word embeddings, including word2vec \cite{Mikolov2013DistributedRO} and GloVe \cite{Pennington2014GloveGV}, also boosts the model performance on the supervised tasks. 

We hypothesize that initializing the deep models with pretrained word embeddings is useful for transferring the knowledge from unsupervised learning to the supervised tasks. We choose to initialize the word embedding matrix in the model with word2vec \cite{Mikolov2013DistributedRO}, GloVe \cite{Pennington2014GloveGV}, and the original method of \cite{Kiros2015SkipThoughtV} that uses random samples from a uniform distribution, respectively. 

\section{Experiment Settings}
\label{settings}
The large corpus that we used for unsupervised training is the BookCorpus dataset \cite{Zhu2015AligningBA}, which contains 74 million sentences from 7000 books in total. 

All of our experiments were conducted in Torch7 \cite{torch}. To make the comparison fair, we follow the encoder design by \cite{Kiros2015SkipThoughtV}. Since the comparison among different recurrent units is not our main focus, we decide to work with GRU, which is fast and stable. In addition, \cite{Chung2014EmpiricalEO} shows that, on language modeling tasks, GRU performs as well as the long short-term memory (LSTM) \cite{Hochreiter1997LongSM}. We also reimplemented the skip-thought model under the same settings, according to \cite{Kiros2015SkipThoughtV}, and the publicly available theano code \footnote{https://github.com/ryankiros/skip-thoughts}. We adopted the multi-GPU training scheme from the Facebook's implementation of ResNet\footnote{https://github.com/facebook/fb.resnet.torch}.

The experiments with bi-directional encoder and unidirectional encoder were both conducted in \cite{Kiros2015SkipThoughtV}, and we follow the design of these experiments. For training efficiency, we didn't follow the exact same dimensionality used. In \cite{Kiros2015SkipThoughtV}, for bi-skip model, the encoder contains a bi-directional GRU with 1200 dimension of each, for uni-skip model, the encoder contains a uni-directional GRU with 2400 dimension, and the decoder is a one-layer with 2400 dimension. 

In our experiments, except for Section \ref{double}, the bi-directional encoder contains a forward and a backward GRU of 300 dimension each, and the uni-directional encoder contains a forward GRU of 600 dimension. After training the 2 models with 2 different encoders separately, we concatenate the vectors produced from 2 encoders to form a sentence representation, and evaluate the performance on evaluation tasks. The decoder is a one-layer unidirectional RNN with GRU, and the dimension is 600. The dimension of word embedding is 300. 

For stable training, we use ADAM \cite{Kingma2014AdamAM} algorithm for optimization. The gradient will be cut off to make it within $[-1,1]$. For the purpose of fast training, all the sentences were zero-padded or clipped to have the same length.

The vocabulary for unsupervised training is set to contain the top 20k most frequent words in BookCorpus. In order to generalize the model trained with relatively small, fixed vocabulary to a large amount of English words, \cite{Kiros2015SkipThoughtV} proposed a word expansion method that learns a linear projection from the pretrained word embeddings word2vec \cite{Mikolov2013DistributedRO} to the learned RNN word embeddings. Thus, the model benefits from the generalization ability of the pretrained word embeddings. 

\begin{table*}[t]
\fontsize{8}{11}\selectfont
\begin{center}
\begin{tabular}{c | c | c c c | c | c c c c c }
\hline
\multirow{2}{*}{Model} & \multirow{2}{*}{WE} & \multicolumn{3}{c|}{SICK} & \multirow{2}{*}{MSRP (Acc/F1)} & \multirow{2}{*}{MR} & \multirow{2}{*}{CR} & \multirow{2}{*}{SUBJ} & \multirow{2}{*}{MPQA} & \multirow{2}{*}{TREC}   \\ \cline{3-5}
 & & $r$ & $\rho$ & MSE &&&&&&  \\ 
 \hline
 \hline
 \multicolumn{11}{c}{\textbf{Plain Connection}} \\
 \hline
bi-T-skip & \multirow{3}{*}{word2vec} & 0.8408 & 0.7649 & 0.2994 & 75.3 / \textbf{83.0} & 76.1 & 80.3 & 92.3 & 87.5 & 86.6 \\
uni-T-skip &	 & 0.8349 & 0.7629 & 0.3084 & 73.7 /	81.9 &	75.7 &	82.1 & 91.3 & 87.4 & 86.4 \\
C-T-skip &	& \textbf{0.8518} & \textbf{0.7808} & \textbf{0.2802} & \textbf{75.7} /	\textbf{83.0} & 76.8 & \textbf{83.2} & \textbf{92.8} & \textbf{88.4} & 87.5 \\
\hline
bi-skip & \multirow{3}{*}{word2vec} & 0.8385 & 0.7618 & 0.3028 & 73.9 / 82.0 & 75.7 & 81.4 & 92.1 &	87.2 & 88.4 \\ 
uni-skip & & 0.8344 & 0.7586 & 0.3098 & 73.6 / 81.6 & 76.2 &	81.8 & 92.2 & 87.6 & 87.0 \\ 
C-skip  & & 0.8492 & 0.7738	& 0.2844 & 74.6 / 82.3 & \textbf{77.0} & 83.0 & 92.7 & 87.9 & \textbf{89.2} \\ 
\hline
\hline
\multicolumn{11}{c}{\textbf{Average+Max Connection}} \\
\hline
bi-T-skip & \multirow{3}{*}{random} & 0.8336 & 0.7612 & 0.3112 & 73.2 /	81.3 & 69.7 & 76.0 & 89.6 &	83.5 & 86.6 \\ 
uni-T-skip &  & 0.8293 & 0.7555 & 0.3180 & 72.5 / 81.0 & 67.3 & 74.9 & 89.0 & 81.1 & 83.6 \\ 
C-T-skip  &  & 0.8458 & 0.7755 & 0.2902 & 74.7 /	82.1 & 70.4 & 76.7 & 90.4 & 83.8 & 84.8 \\ 
\hline
bi-T-skip & \multirow{3}{*}{GloVe} & 0.8444 & 0.7739 & 0.2922 & 75.1 / 82.4 & 74.4 &	79.5 &	90.9 &	85.3 & 87.6 \\ 
uni-T-skip &  & 0.8485 & 0.7711 & 0.2854 & 73.7 / 81.8 & 74.6 & 78.8 & 91.1 &	86.2 & 87.0 \\ 
C-T-skip  &  & 0.8596 & \textbf{0.7903} & 0.2665 & \textbf{75.4} /	\textbf{82.6} & \textbf{75.6} & \textbf{80.4} & 91.9 & 87.0 & 89.0 \\ 
\hline
bi-T-skip &	\multirow{3}{*}{word2vec} & 0.8463 & 0.7744 & 0.2894 &	73.3 /	81.6 &	74.4 & 78.6 & 91.3 & 86.2 & 88.8 \\
uni-T-skip &	 & 0.8466 & 0.7705 & 0.2884 & 74.0 / 81.7 &	73.0 & 78.6 & 91.3 & 85.2 &	88.4 \\
C-T-skip &	 & \textbf{0.8598} & 0.7892	& \textbf{0.2654} &	75.0 /	82.2 &	75.1 &	80.0 &	\textbf{92.2} &	\textbf{87.2} & \textbf{90.0} \\
\hline
\hline
\multicolumn{11}{c}{\textbf{Doubled Encoder's Dimension vs. Results reported by \cite{Kiros2015SkipThoughtV}}} \\
\hline
bi-T-skip &	\multirow{3}{*}{word2vec} & 0.8503 & 0.7796 & 0.2823 &	74.4 /	82.2 &	74.8 & 80.3 & 91.8 & 87.0 & 88.2 \\
uni-T-skip &	 & 0.8486 & 0.7784 & 0.2857 & 74.3 / 82.4 &	72.9 &	78.0 &	90.7 &	85.7 &	86.4 \\
C-T-skip &	 & \textbf{0.8611} & \textbf{0.7946} & \textbf{0.2634} & \textbf{74.5} / \textbf{82.2} & 75.4 &	 \textbf{80.3}  & 92.2 &  \textbf{87.4} & 88.4 \\
\hline
bi-skip \cite{Kiros2015SkipThoughtV} &	\multirow{3}{*}{random} & 0.8405 &	0.7696 &	0.2995 &	71.2 / 81.2 &	73.9 & 77.9  &	92.5	& 83.3 &	89.4 \\
uni-skip \cite{Kiros2015SkipThoughtV} &	 &  0.8477 &	0.7780 &	0.2872 &	73.0 / 81.9 &	75.5 &	79.3   &	92.1	& 86.9	& 91.4  \\
C-skip \cite{Kiros2015SkipThoughtV} &	 & 0.8584 &	0.7916 &	0.2687 &	73.0 / 82.0 &	\textbf{76.5} & 80.1 & \textbf{93.6} & 87.1  & \textbf{92.2} \\
\hline
\end{tabular}
\end{center}
\caption{The model name is given by \emph{encoder type - model type}. Bold numbers indicate the best results among the models in each section. Our trimmed skip-thought models slightly outperform the skip-thought models. The model with doubled-sized encoder has average+max connection.}
\label{quantitative}
\end{table*}

\section{Quantitative Evaluation}
\label{QE}
We compared our proposed trimmed skip-thought model with the skip-thought model on 7 evaluation tasks, which include semantic relatedness (SICK) \cite{Marelli2014ASC}, paraphrase detection (MSRP) \cite{Dolan2004UnsupervisedCO}, question-type classification (TREC) \cite{Li2002LearningQC}, and 4 benchmark sentiment and subjective datasets, which includes movie review sentiment (MR) \cite{Pang2005SeeingSE}, customer product reviews (CR) \cite{Hu2004MiningAS}, subjectivity/objectivity classification (SUBJ) \cite{Pang2004ASE}, and opinion polarity (MPQA) \cite{Wiebe2005AnnotatingEO}. After unsupervised training on the BookCorpus dataset, we fix the parameters in the encoder, and apply it as a sentence representation extractor on the 7 tasks. 

For SICK and MSRP tasks, we adopt the feature engineering idea proposed by \cite{Tai2015ImprovedSR}. For a given sentence pair, the encoder computes a pair of representations, denoted as $u$ and $v$, and the concatenation of the component-wise product $u\cdot v$ and the absolute difference $|u-v|$ is regarded as the feature vector for the given sentence pair. Then we train logistic regression on the feature vector to predict the semantic relatedness score. The evaluation metrics for SICK are Pearson’s $r$, Spearman’s $\rho$, and mean squared error $MSE$, and the performance is reported as accuracy and F1-score (Acc/F1) for MSRP. The performance on TREC is presented as test accuracy, and 10-fold cross validation is applied to evaluate the model on the MR, CR, SUBJ, and MPQA classification benchmarks.

Table \ref{quantitative} presents the models and results, where the model name is given by \emph{encoder type - model type}. Three types of  encoder are denoted as \emph{uni-}, \emph{bi-}, and \emph{C-} in Table \ref{quantitative}, and the \emph{C-} refers to the concatenation of 2 vector representations computed from \emph{uni-}encoder model and \emph{bi-}encoder model. \emph{-T-skip} refers to our trimmed skip-thought model, and \emph{-skip} refers to the skip-thought model.

\subsection{Trimmed skip-thought vs. Skip-thought}
We first compare our trimmed skip-thought model with our implemented skip-thought model, to check the neighborhood hypothesis. In this comparison, all the models use the plain connection, which means that, the sentence representation is the hidden state at the last time step. 

Table \ref{quantitative} presents the results of 3 trimmed skip-thought models, and 3 of our implemented skip-thought models. From the table, we can tell that our trimmed skip-thought models perform slightly better than the skip-thought models overall, yet not significantly, but the performance on the TREC dataset is worse than the skip-thought models. The general performance comparison between our trimmed skip-thought model and the skip-thought model proves that the neighborhood hypothesis is reasonable, which means that the neighborhood information is effective for distributed sentence representation learning. In addition, our trimmed skip-thought model runs faster in training, since our model only needs to reconstruct its next sentence while the skip-thought model needs to reconstruct its two surrounding sentences.

Unlike the results in \cite{Tang2017Rethinking}, these models presented in this paper contain word embeddings with lower dimension, which is half of that in \cite{Tang2017Rethinking}, and GRUs with much smaller size. Also, our models presented here use word2vec \cite{Mikolov2013DistributedRO} as word embeddings initialization, which is different from random initialization applied in \cite{Tang2017Rethinking}.

The results of our implemented skip-thought model differ from those presented in \cite{Kiros2015SkipThoughtV}, (also presented here in the last section in Table \ref{quantitative}), since our implementation contains much fewer parameters than the original skip-thought model, and it has word2vec \cite{Mikolov2013DistributedRO} initialization. Overall, our implementation reaches similar performance on all tasks except Sick. The comparison with the original skip-thought model shows that our implementation of skip-thought model is reasonable.

\subsection{Plain Connection vs. Average+Max Connection}
We further compare the effect of two different connections between the encoder and the decoder. The results are presented in Table \ref{quantitative}. As we expected, our proposed trimmed skip-thought model benefits from the \emph{Average+Max Connection} on judging the relationship of a sentence pair. The performance on SICK task get improved. However, the performance on 2 classification benchmarks, MR and CR, slightly drops, compared to our model with plain connection. The overall performance on the evaluation tasks reaches the results reported in \cite{Kiros2015SkipThoughtV} except TREC, which shows that our model with \emph{Average+Max Connection} could be a fast, lighter-weight alternative to the skip-thought model. 
See Table \ref{params} for detailed parameter counts.

\begin{table}[h]
\fontsize{9}{12}\selectfont
\begin{center}
\begin{tabular}{c | c | c | c }
\hline
Model & RNNs & Embedding & Prediction\\
\hline
\hline
uni-T-skip (ours) & 4.32M & \multirow{4}{*}{6M} & \multirow{4}{*}{12M} \\ \cline{1-2}
bi-T-skip (ours) & 3.24M & & \\ \cline{1-2}
uni-T-skip-double (ours) & 10.8M & & \\ \cline{1-2}
bi-T-skip-double (ours) & 6.48M & & \\
\hline
\hline
uni-skip \cite{Kiros2015SkipThoughtV} & 69.12M & \multirow{2}{*}{12.4M} & \multirow{2}{*}{48M} \\ \cline{1-2}
bi-skip \cite{Kiros2015SkipThoughtV} & 51.84M & & \\
\hline
\end{tabular}
\end{center}
\caption{The table presents the number of parameters in each part of model. \emph{RNNs}, \emph{Embedding}  and \emph{Prediction} refer to the recurrent parts, the word embedding, and the linear prediction layer in model. \emph{-double} means the encoder with GRU with doubled dimension. Our models have much fewer parameters than the skip-thought models, even with the double-sized encoder.}
\label{params}
\end{table}

\subsection{Word Embedding Initialization}
The second section in Table \ref{quantitative} presents the comparison among 3 different initializations. After training, we learn a linear mapping from the word2vec \footnote{https://code.google.com/archive/p/word2vec/} embedding space to RNN word embedding space, regardless of what initialization was applied in the model as in \cite{Kiros2015SkipThoughtV}. 

Generally, the models with pretrained word embeddings as initialization perform better on the evaluation tasks than those with random initialization, which shows that a good initialization for word embeddings helps the model to better transfer knowledge from unsupervised training.

One thing worth mentioning here, for the models initialized with GloVe \footnote{https://nlp.stanford.edu/projects/GloVe/}, we also trained a linear projection from GloVe word embeddings to the RNN word embeddings. The performance on SICK and MSRP is as good as other models presented in Table \ref{quantitative}, but the word expansion from GloVe embeddings gave bad performance on 5 classification benchmarks, so we didn't include the results. 

\subsection{Doubling Encoder's Dimension}
\label{double}
In our experiments above, the encoder is either a bi-directional GRU with 300 dimension each or a uni-directional GRU with 600 dimension. With the average+max connection, the dimension of a sentence representation is 1200. We hypothesized that a model with larger encoder size could also improve the performance on evaluation tasks. Hence, we double the dimension of the encoder, which is now either a bi-GRU with 600 dimension each or a uni-GRU with 1200 dimension. As a result, the sentence representation is a 2400-dimension vector, which matches the dimensionality of the representation reported in \cite{Kiros2015SkipThoughtV}. Table \ref{quantitative} represents the results.

Our trimmed skip-thought models with doubled encoder performs better than the skip-thought models report in \cite{Kiros2015SkipThoughtV} on SICK and MSRP, and have comparable results on 4 classification benchmarks. The performance is worse than the original skip-thought model on TREC. The training time and inference time are significantly less than that for the original skip-thought model. The cut down on the training time comes from the neighborhood hypothesis\cite{Tang2017Rethinking} and many fewer parameters in the model. 

\begin{table*}[t]
\fontsize{9}{12}\selectfont
\begin{center}
\begin{tabular}{l}
\hline
\hline
\textbf{i wish i had a better answer to that question . }\\
i had no good answer to that question . \\
i had n't really wanted an answer to that particular question . \\
\hline
\textbf{do you want me to meet you ?}\\
alright , where do you want me to meet you ?\\
where can i meet you ?\\
\hline
\textbf{my phone buzzed and i awoke from my trance .} \\
my cell phone ringing woke me up with a jolt . \\
my cell phone buzzed in my lap , and i looked down at my text message . \\
\hline
\textbf{my heart was racing so fast that it might explode right out of my chest . } \\
my heart was pounding so hard it felt as if it might jump out of my chest . \\
my heart felt like it was going to explode out of my chest . \\
\hline
\textbf{i threw my bag on my bed and took off my shoes .}\\
i sat down on my own bed and kicked off my shoes . \\
i fell in bed without bothering to remove my shoes . \\
\hline
\hline
\end{tabular}
\end{center}
\caption{In each section, the first sentence is the query, the second one is the nearest neighbor retrieved from the database, and the third one is the 2nd nearest neighbor. The similarity between every sentence pair is measure by the cosine similarity in the representation space.}
\label{sr}
\end{table*}

\section{Qualitative Investigation}
\label{QI}
We conduct investigation on our trimmed skip-thought model qualitatively. The model being studied here contains bi-GRU as encoder with 300 dimension of each, one-layer GRU as decoder with 600 dimension, and average+max connection.
\subsection{Sentence Retrieval}
For this task, 1000 sentences were selected as the query set, and 1 million sentences were picked up as the database. All the sentences come from the training corpus. Cosine distance is applied to measure the distance in the representation space. See Table \ref{sr} for several samples. Most of retrieved sentences look semantically related and can be viewed as the sentential contextual extension to the query sentences.

\subsection{Conditional Sentence Generation}
\begin{table}[!h]
\fontsize{10}{12}\selectfont
\begin{center}
\begin{tabular}{l}
\hline
\hline
i 'm not going to let you go . \\
i 'm not sure if i should be mad at him or not . \\
i did n't want to hear it . \\
i 'm not sure i would ever be able to get her to agree with me . \\
`` i 'm not going to be a little girl , '' i said . \\
i 'm not sure i could ever be with him . \\
i was n't sure if i was going to be a part of the night or the next day . \\
\hline
\hline
\end{tabular}
\end{center}
\caption{Samples of the generated sentences.}
\label{sents}
\end{table}
The decoder in our trimmed skip-thought model was trained in language modeling fashion, it is reasonable to analyze the sentences generated from the decoder after training. Since the sentence generation process is conditional on the representations produced from the encoder, we first randomly pick up sentences from the training corpus, and forward the model to get the output from the decoder for each of them. Greedy decoding is applied for sentence generation. Table \ref{sents} presents the generated sentences. 

We observe that, the generated sentences tend to start with \emph{i 'm not}, and \emph{i do n't}. It might be caused the corpus bias, since there exists a large amount of sentences that start with \emph{i 'm not}, \emph{i do n't}, etc. In addition, the decoder is trained to reconstruct the next sentence in the model, which could be think of as a sentential contextual extension of the input sentence, while the generated sentences rarely are related to the associated input sentences, which is same for the skip-thought models. More investigations are needed for the conditional sentence generation.

\section{Related Work}
Previously, \cite{Mikolov2013EfficientEO} proposed the continuous bag-of-words (CBOW) model and the skip-gram model for distributed word representation learning. The main idea is learn a word representation by discovering the context information from the surrounding words. \cite{Mikolov2013DistributedRO} improved the skip-gram model, and empirically showed that additive composition of the learned word representations successfully captures contextual information of phrases and sentences, which is a strong baseline model for NLP tasks. Similarly, \cite{Le2014DistributedRO} proposed a method to learn a fixed-dimension vector for each sentence by predicting the words within the given sentence. However, after training, the representation for a new sentence is hard to derive, since it requires optimizing the sentence representation towards an objective.

Recent research in deep learning shows that, the word representation and the their composition could be done at the same time in an end-to-end machine learning system. LSTM-based autoencoder model for language representation learning was proposed by \cite{Dai2015SemisupervisedSL}. For a specific dataset, the model first was trained in an unsupervised fashion and then finetuned for the supervised task. The model didn't outperform previous CBOW models significantly, but it shows that knowledge learned through unsupervised pretraining could be transfered to augment the performance on the supervised tasks.

The skip-thought model was proposed by \cite{Kiros2015SkipThoughtV} for learning a generic, distributed sentence encoder, and its key idea was inspired by the skip-gram model \cite{Mikolov2013EfficientEO}. The results on 8 evaluation tasks are promising with no finetuning on the encoder, and some of the results reach other supervised trained models. In \cite{Triantafillou2016TowardsGS}, they finetuned the skip-thought models on the SNLI corpus \cite{Bowman2015ALA}, which shows that the skip-thought pretraining scheme is generalizable to other specific NLP tasks.

\cite{Hill2016LearningDR} pointed out that the skip-thought model made use of the sentence-level distributional hypothesis \cite{harris1954distributional,Polajnar2015AnEO}. Following the same hypothesis, \cite{Hill2016LearningDR} proposed FastSent model. It takes summation of the word embeddings in a sentence as the sentence representation, and predicts the words in both the previous sentence and the next sentence. It is an simplification of the skip-thought model, which assume the composition function of the words is summation. The results on SICK is comparable with the skip-thought model, while the skip-thought model still outperforms the FastSent model on the other six evaluation tasks. Later, Siamese CBOW \cite{Kenter2016SiameseCO} aimed to learn the word representations to make the cosine similarity of adjacent sentences in the representation space larger than that of  sentences which are not adjacent. The reported comparison with the skip-thought and FastSent models on SICK dataset was convincing that the Siamese CBOW captures better sentence semantics, while no other comparisons on evaluation tasks were reported.

Instead of learning to reconstruct the sentences which are adjacent to the current sentence, \cite{jernite2017discourse} proposed a model that learns to categorize the manually defined relationships of two input sentences. The model encodes two sentences in two representations, respectively, and the classifier on top of the representations judges 1) whether the two sentences are adjacent to each other, 2) whether the two sentences are in the correct order, and 3) whether the second sentence starts with a conjunction phrase. The proposed model runs faster than the skip-thought model, since it only contains an encoder and no decoder is required. However, only the result on microsoft paraphrase detection task is similar to that of the skip-thought model, and the results on other tasks are not as good.
\vspace{-0.2cm}

\section{Conclusion} 
We proposed 3 techniques for trimming and also improving the skip-thought model\cite{Kiros2015SkipThoughtV}, which includes dropping off one decoder, incorporating non-linear non-parametric connection, and initializing with pretrained word vectors. We empirically showed the effectiveness of our proposed techniques. In addition, our proposed trimmed skip-thought model contains much fewer parameters, which runs much faster than skip-thought model. Furthermore, our model could be facilitated by various connection methods between the encoder and the decoder, and benefit from a larger model size. Future research could make use of proposed techniques on unsupervised representation learning, and generalize to more sophisticated model types.

\section*{Acknowledgments}
We gratefully thank Jeffrey L. Elman, Benjamin K. Bergen, and Seana Coulson for insightful discussion, and thank Thomas Donoghue, and Reina Mizrahi for suggestive chatting. We also thank Adobe Research Lab for GPUs support, and thank NVIDIA for DGX-1 trial as well as support from NSF IIS 1528214 and NSF SMA 1041755.

\bibliography{nips_2017.bib}

\end{document}